\documentclass{article}
\usepackage{spconf,amsmath,graphicx}
\usepackage{amsfonts}
\usepackage{amsmath,bm}
\usepackage{adjustbox}
\usepackage{algorithmicx}
\usepackage{algpseudocode}
\usepackage[ruled]{algorithm}
\usepackage{graphicx}
\usepackage{caption}
\usepackage{graphicx,times,amsmath} 
\usepackage{caption}
\usepackage{subcaption}
\usepackage[rightcaption]{sidecap}
\usepackage{caption}
\usepackage{multirow}
\usepackage{hhline}
\usepackage{amsmath}
\usepackage{epstopdf}
\usepackage{tabularx}
\usepackage{tikz}
\usetikzlibrary{matrix}
\usepackage{mathtools, nccmath}
\usepackage{pst-node}

\graphicspath{ {images/} }

\usepackage{alphalph}

\title{Pruning of Convolutional Neural Networks Using Ising Energy Model}

\name{Hojjat Salehinejad, Member, IEEE, and Shahrokh Valaee, Fellow, IEEE}
\address{Department of Electrical \& Computer Engineering, University of Toronto, Toronto, Canada \\
\textit{hojjat.salehinejad@mail.utoronto.ca, valaee@ece.utoronto.ca}}

\usepackage{fancyhdr}
\fancypagestyle{pageStyleOne}{%
\normalfont\footnotesize
    \fancyhf{}
\fancyhead[L]{\fontsize{8}{8} \selectfont This paper is accepted for presentation at IEEE International Conference on Acoustics, Speech and Signal Processing (IEEE ICASSP), 2021.}
}

\begin{document}
\newcommand*{\img}{%
  \includegraphics[
    width=\linewidth,
    height=20pt,
    keepaspectratio=false,
  ]{example-image-a}%
}

\maketitle
    \thispagestyle{pageStyleOne}

\begin{abstract}
 Pruning is one of the major methods to compress deep neural networks. In this paper, we propose an Ising energy model within an optimization framework for pruning convolutional kernels and hidden units. This model is designed to reduce redundancy between weight kernels and detect inactive kernels/hidden units. 
Our  experiments using ResNets, AlexNet, and SqueezeNet on CIFAR-10 and CIFAR-100 datasets show that the proposed method on average can achieve a  pruning rate of more than $50\%$ of the trainable parameters with approximately $<10\%$ and $<5\%$ drop of Top-1 and Top-5 classification accuracy, respectively. 

\end{abstract}
\section{Introduction}
\label{sec:intro}
 Deployment of deep neural networks (DNNs) in inference mode is challenging for applications with limited resources such as in edge devices~\cite{lin2019toward, salehinejad2019ising}. Pruning is one of the major approaches for compressing a DNN by permanently dropping a subset of network parameters. Pruning methods are divided into unstructured and structured approaches. Unstructured pruning does not follow a specific geometry and removes any subset of the weights~\cite{anwar2017structured}. Structured pruning typically follows a geometric structure and happens at channel, kernel, and intra-kernel levels~\cite{anwar2017structured, cheng2017survey}. One of the early attempts for pruning neural networks was to use second derivative information to minimize a cost function that reduces network complexity by removing excess number of trainable parameters and further training the remaining of the network to increase inference accuracy~\cite{lecun1990optimal}. \textit{Deep Compression} is a popular pruning method which has three stages that are pruning, quantization, and Huffman coding~\cite{han2015deep}. This method works by pruning all connections with weights below a threshold followed by retraining the sparsified network. 

We have proposed an Ising energy model in~\cite{salehinejad2019ising} for pruning hidden units in multi-layer perceptron (MLP) networks, \cite{salehinejad2019ising2}. In this paper, we propose IPruning, which targets pruning convolutional kernels, including all corresponding input/output connections, and hidden units based on modeling a DNN as a graph and quantifying interdependencies among trainable variables using the Ising energy model.
 A DNN is modeled as a graph where the nodes represent kernels/hidden units, and edges represent the relationship between nodes. This relationship is modeled using entropy of feature maps between convolutional layers and relative entropy (Kullback–Leibler (KL) divergence) between convolutional kernels in a layer. These relationships are represented as weights in an Ising energy model, which targets dropping kernels with low activity and eliminating redundant kernels. We initiate a set of candidate pruning state vectors which correspond to different subgraphs of the original graph. The objective is to search for the state vector that minimizes the Ising energy of the graph. Each step of the optimization procedure happens within a training iteration of the network, where only the kernels identified by the best pruning state vector are trained using backpropagation. This is indeed similar to training with dropout~\cite{labach2019survey}, where the original network is partially trained. However, after a number of iterations the set of candidate state vectors can converge to a best pruning state vector, which represents the pruned network\footnote{The codes and more details of experiments setup is available at: \textit{https://github.com/sparsifai/ipruning}}.

\section{Proposed Ising Pruning Method}

The weights of a DCNN with $L$ layers are defined as the set  $\bm{\Theta}=\{\Theta^{[1]},...,\Theta^{[L]}\}$, where $\Theta^{[l]}\in\bm{\Theta}$ is the set of weights in layer $l$, $\Theta_{i}^{[l]}$ is the weight  kernel $i$ of size $N_{i}^{[l]}$, and $N^{[l]}$ is the number of weight kernels in the convolution layer $l$. Similarly, in a dense layer $N^{[l]}$ is the number of weights from layer $l$ to the next layer $l+1$. Generally, a feature map is constructed using the convolution operation defined as
\begin{equation}
 \begin{split}
  \mathbf{F}^{[l]}_{i}&= \sigma(\Theta^{[l]}_{i}\star\mathbf{F}^{[l-1]}),
  \end{split}
\end{equation}
where $\mathbf{F}^{[l]}_{i}$ is the feature map $i$ in layer $l$, $\mathbf{F}^{[l-1]}$ is the set of feature maps from the previous layer, $\sigma(\cdot)$ is the activation function, and $\star$ is the convolution operation. 
The key questions is ``\textit{How to detect redundant and inactive kernels in a DCNN?}''. To answer this question, we suggest to quantitatively evaluate activity and redundancy of the kernels using entropy and KL divergence as follows.

\subsection{Measuring Kernels Activity}
Feature maps are the activation values of a convolutional layer, representing activation of weight kernel for a given input. 
We use feature maps of a kernel as a means of evaluating its activation. Assuming $\sigma(x)=max(0,x)$, a feature map value $f_{i,j}\in \mathbb{R}_{\geq 0}$, where $j$ is the $j^{th}$ element of $\mathbf{F}_{i}$, is generally a real number in a continuous domain. We quantify a feature map in a discrete domain $\Lambda$ by mapping the feature map values as $\mathbb{R}_{\geq 0} \xrightarrow{\mathcal{Q}(\cdot)}\Lambda$ where  $\Lambda=\{0,1,...,255\}$ is an 8-bit discrete state space and 
\begin{equation}
\mathcal{Q}(f_{i,j})=\lfloor 255\cdot \frac{f_{i,j}}{max(\mathbf{F}_{i})}\rceil,
\end{equation}
where $\lfloor\cdot\rceil$ is the round to the nearest integer function. Let us define the random variable $F =\mathcal{Q}(f_{i,j})$ with possible outcome $\lambda\in\Lambda$. Then, the probability of $\lambda$ is
\begin{equation}
p_{F}(\lambda)=n_{\lambda}/|\mathbf{F}_{i}| \quad \forall \quad  \lambda\in\Lambda,
\label{eq:pmf_fms}
\end{equation}
where $n_{\lambda}$ is the number of times $\lambda$ occurs and $|\cdot|$ is the cardinality of the feature map. The entropy of the feature map $\mathbf{F}_{i}$ is then defined as
\begin{equation}
\mathcal{H}_{i}=-\sum_{\lambda\in\Lambda}p_{F}(\lambda)log_{2}p_{F}(\lambda).
\label{eq:entropy}
\end{equation}

\subsection{Measuring Kernels Redundancy}
\label{sec:kl}
The weights in neural networks are generally initialized from a normal distribution.
A DCNN can have redundancy between kernels in a layer. Removing the redundant kernels prunes the network while may slightly drop the classification accuracy. A kernel $\Theta$ is generally a three-dimensional tensor of size $K_{1}\times K_{2}\times N$ where $K=K_{1}\times K_{2}$ is the size of a filter and $N$ is the number of filters, corresponding to the number of input channels. Therefore, we can represent the weights in a kernel with $K$ sets which are ${W_{1},...,W_{K}}$ 
where ${W_{k}=\{\theta_{k,1},...,\theta_{k,N}\}}$ and $k\in\{1,...,K\}$. Let us assume the weights $W_{k}$ have a normal distribution. Hence, for the kernel $i$ we have a multivariate normal distribution with means ${\bm{\mu}_{i}=(\mu_{i,1},...,\mu_{i,K})}$ and the ${K\times K}$ covariance matrix ${\bm{\Sigma}}_{i}$. The distributions $\mathcal{N}_{i}(\bm{\mu}_{i},\bm{\Sigma}_{i})$ and $\mathcal{N}_{j}(\bm{\mu}_{j},\bm{\Sigma}_{j})$ of two given kernels $i$ and $j$, respectively, have the same dimension. Hence, we can compute the KL divergence between the two kernels $i$ and $j$ as
\begin{equation}
\begin{split}
\mathcal{D}_{KL}(\mathcal{N}_{i}||\mathcal{N}_{j})=\\&
\frac{1}{2}\Big(
tr(\bm{\Sigma}_{j}^{-1}\bm{\Sigma}_{i}) 
+ (\bm{\mu}_{j}-\bm{\mu}_{i})^\top \bm{\Sigma}_{j}^{-1} (\bm{\mu}_{j}-\bm{\mu}_{i}))\\
&-K
+ln\Big(\frac{|\bm{\Sigma}_{j}|}{|\bm{\Sigma}_{i}|}\Big)
\Big),
\end{split}
\label{eq:kldivergence}    
\end{equation}
where $tr(\cdot)$ is the trace and $|\cdot|$ is the determinant.

\subsection{Ising Energy Formulation}
A neural network $\mathcal{F}$ has the set of layers $A=\{A_{1}\cup A_{2}\}$, where $A_{1}$ and $A_{2}$ are the set of convolutional and dense layers, respectively. Obviously, the sets $A_{1}$ and $A_{2}$ are disjoint (i.e. $A_{1}\cap A_{2}=\emptyset$). Hereafter we refer to a hidden unit or a convolutional kernel a unit for simplicity. A binary state vector $\mathbf{s}$ with length $D$ represents the state of the units, where $s_{d}\in\{0,1\}\:\forall\:d\in\{1,...,D\}$. If $s_{d}=0$ unit $d$ is inactive and if $s_{d}=1$ the unit participates in training and inference. Therefore, the state vector $\mathbf{s}$ represents a subnetwork of the original network. The unit $d$ belongs to a layer $l\in A$.

Let us represent the network $\mathcal{F}$ as a graph $G = (\mathcal{D},\Gamma)$, where $\mathcal{D}$ is the set of vertices
(nodes) with cardinality $D$ and $\Gamma$ is the set of edges (connections) with weight $\gamma_{d,d'}$ between vertices $d$ and $d'$. The graph has two types of connections, where the connection between vertices of a layer is bidirectional and the connection between layers is unidirectional. In dense layers, unidirectional connections exist between nodes of two layers where each node has a state $s_{d}\in\{0,1\}$, except in the last layer (logits layer), where $s_{d}=1$. We are interested in pruning  the vertices and all corresponding edges. 

We model the dependencies between vertices in the graph $G$ using the Ising energy model as
\begin{equation}
\mathcal{E}=-\sum_{d\in \mathcal{D}}\sum_{d'\in \mathcal{D}}\gamma_{d,d'}s_{d}s_{d'} - b\sum_{d\in \mathcal{D}}s_{d},
\label{eq:ising}    
\end{equation}
where $b$ is the bias coefficient and $\gamma_{d,d'}$ is the weight between the vertices $d$ and $d'$ defined as
\begin{equation}
\gamma_{d,d'}=\begin{cases}
           \mathcal{D}_{KL}(\mathcal{N}_{d}||\mathcal{N}_{d'})-1\:$if$\: d,d'\in l \: \& \:l\in A_{1} \\
           \mathcal{H}_{d}-1\:$if$\: d\in l,d'\in l+1 \: \& \:l,l+1\in A_{1}\\
           \mathcal{A}_{d}-1\:$if$\: d\in l,d'\in l+1 \: \& \:l,l+1\in A_{2}\\
           0 \:\:\:\:$otherwise$\\
        \end{cases}, 
        \label{eq:cases}
\end{equation}
where $\mathcal{H}_{d}$ is calculated using~(\ref{eq:entropy}). Similar to the approach we have proposed in~\cite{salehinejad2019ising} for hidden units in dense layer, we have
\begin{equation}
\mathcal{A}_{i}=tanh(a_{i}),
\label{eq:denseent}
\end{equation} 
which maps the activation value $a_{i}$ of the unit $i$, generated by the ReLU activation function such that a dead unit has the lowest $\mathcal{A}_{i}$ and a highly activated unit has a high $\mathcal{A}_{i}$.  
In~(\ref{eq:cases}) a high weight is allocated to the unidirectional connections of a unit with high activation value and a high weight is allocated to the bidirectional connections with high KL divergence, and vice-versa. From another perspective, the first case allocates small weight to low-active units and the latter case allocates small weight to redundant units. 

Assuming all the states are active (i.e. $s_{d}=1\;\forall\;d\in\mathcal{D}$), the bias coefficient is defined to balance the interaction term and the bias term by setting $\mathcal{E}=0$. Hence, 
\begin{equation}
\begin{split}
     b &= -\frac{\sum_{d\in \mathcal{D}}\sum_{d'\in \mathcal{D}}\gamma_{d,d'}s_{d}s_{d'}}{\sum_{d\in \mathcal{D}}s_{d}}\\
     &=-\frac{|\bm{\gamma}|}{D},
\end{split}
\end{equation}
where $|\bm{\gamma}|$ is the sum of weights $\bm{\gamma}$ and $\sum_{d\in \mathcal{D}}s_{d}=D$. Minimizing (\ref{eq:ising}) is equivalent to finding a state vector which represents a sub-network of $\mathcal{F}$ with a smaller number of redundant kernels and inactive units.

\begin{algorithm}[t]
\small
\begin{algorithmic} 
\State Set $t$ = 0 // Optimization counter
\State Initiate the neural network $\mathcal{F}$
\State Set $\mathbf{S}^{(0)}\sim Bernoulli(P=0.5)$ // States initialization
\State Set $\Delta\mathbf{s}\neq 0$ // Early state threshold
\For{ $ \mathit{i_{epoch}} = 1 \rightarrow \mathit{N_{epoch}}$} // Epoch counter
\For{ $ \mathit{i_{batch}} = 1 \rightarrow \mathit{N_{batch}}$} // Batch counter
\State $t = t+1$
\If{$\Delta\mathbf{s}\neq 0$}
\If{$i_{epoch}=1 \; \& \; i_{batch}=1$}
\State Compute energy of $\mathbf{S}^{(0)}$ using~(\ref{eq:ising})
\EndIf

\For {$i=1\rightarrow S$} // States counter
\State Generate mutually different $i_{1},i_{2},i_{3}\in \{1,...,S\}$
\For {$d=1\rightarrow D$} // State dimension counter
\State Generate a random number $r_{d}\in[0,1]$
\State Compute mutation vector $v_{i,d}$ using (\ref{eq:mutation})
\State Compute candidate state $\tilde{s}^{(t)}$ using (\ref{eq:crossover})
\EndFor
\EndFor
\State Compute energy loss of $\tilde{\mathbf{S}}^{(t)}$ using~(\ref{eq:ising})
\State Select $\mathbf{S}^{(t)}$ and corresponding energy using (\ref{eq:selection})
\State Select the state with the lowest energy from $\mathbf{S}^{(t)}$ as $\mathbf{s}^{(t)}_{b}$
\Else
\State $\mathbf{s}^{(t)}_{b}=\mathbf{s}^{(t-1)}_{b}$
\EndIf
\State Temporarily drop weights of $\mathcal{F}$ according to $\mathbf{s}^{(t)}_{b}$
\State Compute cross-entropy loss of the sparsified network
\State Perform backpropagation to update active weights
\EndFor
\State Update $\Delta\mathbf{s}$ for early state convergence using (\ref{eq:stateconvergence})
\EndFor
\end{algorithmic}
\small
  \caption{IPruning}
  \label{alg:IsingEnergy-basedDropout}
\end{algorithm}

\subsection{Optimization of Ising Energy}
Algorithm~\ref{alg:IsingEnergy-basedDropout} shows different steps of IPruning. The process of searching for the pruning state vector with lowest energy is incorporated into the typical training of the neural network $\mathcal{F}$ with backpropagation. First, a population of candidate state vectors is initiated and then the Ising energy loss is computed for each vector. Then, the population of vectors is evolved on the optimization landscape of states with respect to the Ising energy and the state with lowest energy is selected. Dropout is performed according to the selected state vector and only active weights are updated with backpropagation. The population is then evolved and the same procedure is repeated until the population of states converges to a best state solution or a predefined number of iterations is reached.

Let us initialize a population of candidate states $\mathbf{S}^{(t)}\in\mathbb{Z}_{2}^{S\times D}$ such that $\mathbf{s}^{(t)}_{i}\in\mathbf{S}^{(t)}$, where $t$ is the iteration and $s_{i,d}^{(0)}\sim Bernoulli(P=0.5)$ for ${i\in\{1,...,S\}}$ and ${d\in\{1,...,D\}}$. A state vector $\mathbf{s}^{(t)}_{j}\in\mathbf{S}^{(t)}$ selects a subset of the graph $G$.

The optimization procedure has three phases which are mutation, crossover, and selection. Given the population of states $\mathbf{S}^{(t-1)}$, a mutation vector is defined for each candidate state $\mathbf{s}_{i}^{(t-1)}\in\mathbf{S}^{(t-1)}$ as 
\begin{equation}
v_{i,d}=\begin{cases}
               1-s_{i_{1},d}^{(t-1)} \:\:\:\:$if$\:\:\:s_{i_{2},d}^{(t-1)}\neq s_{i_{3},d} ^{(t-1)}\;$\&$\; r_{d}<F\\
               s_{i_{1},d}^{(t-1)} \:\:\:\:\:\:\:\:\:\:\:$ otherwise $
            \end{cases},
\label{eq:mutation}
\end{equation}
for $d\in\{1,..,D\}$ where $i_{1},i_{2},i_{3}\in \{1,...,S\}$ are mutually different, $F$ is the mutation factor~\cite{salehinejad2017micro}, and $r_{d}\in[0,1]$ is a random number. 
The next step is to crossover the mutation vectors to generate new candidate state vectors as
\begin{equation}
\tilde{s}^{(t)}_{i,d}=\begin{cases}
               v_{i,d} \:\:\:\:\:\:\:\:\:\:\:\:$if$\:\:\: r'_{d}\in[0,1] \leq C\\
               s_{i,d}^{(t-1)} \:\:\:\:\:\:\:\:\:\:\:$ otherwise $
            \end{cases},
\label{eq:crossover}
\end{equation}
where $C=0.5$ is the crossover coefficient~\cite{salehinejad2017micro}. The parameters $C$ and $F$ control exploration and exploitation of the optimization landscape. Each generated state $ \tilde{\mathbf{s}}_{i}^{(t)}$ is then compared with its corresponding parent with respect to its energy value $\tilde{\mathcal{E}}^{(t)}_{i}$ and the state with smaller energy is selected as
\begin{equation}
\mathbf{s}_{i}^{(t)}=\begin{cases}
               \tilde{\mathbf{s}}_{i}^{(t)} \:\:\:\:\:\:\:\:\:\:$if$\:\:\:    \tilde{\mathcal{E}}^{(t)}_{i}\leq \mathcal{E}^{(t-1)}_{i} \\
              \mathbf{s}_{i}^{(t-1)} \:\:\:\:\:$ otherwise $
            \end{cases}\:\forall\:i\in\{1,...,S\}.
\label{eq:selection}
\end{equation}
The state with minimum energy $\mathcal{E}_{b}^{(t)}=min\{\mathcal{E}_{1}^{(t)},...,\mathcal{E}_{S}^{(t)}\}$ is selected as the best state $\mathbf{s}_{b}$, which represents the sub-network for next training batch. This optimization strategy is simple and feasible to implement in parallel for a large $S$. 

After a number of iterations, depending on the capacity of the neural network and complexity of the dataset, all the states in $\mathbf{S}^{(t)}$ may converge to the best state vector  $\mathbf{s}_{b}\in\mathbf{S}^{(t)}$ with the Ising energy $\mathcal{E}_{b}^{(t)}$. Hence, we can define
\begin{equation} 
\Delta\mathbf{s} =
\mathcal{E}_{b}^{(t)} -
\frac{1}{S}\sum\limits_{j=1}^{S}\mathcal{E}_{j}^{(t)},
\label{eq:stateconvergence}
\end{equation}
such that if $\Delta\mathbf{s}=0$, we can call for an {early state convergence} and continue training by fine-tuning the sub-network identified by the state vector $\mathbf{s}_{b}$.

\begin{table}[!ht]
\captionsetup{font=footnotesize}

\caption{Classification performance on the test datasets. $R$ is kept trainable parameters and $\#p$ is approximate number of trainable parameters. All the values except loss and $\#p$ are in percentage. (F) refers to full network used for inference and (P) refers to pruned network using \textit{IPruning}.}

\begin{subtable}{0.99\linewidth}
\centering
\captionsetup{font=footnotesize}
\caption{ \textbf{CIFAR-10} }
\begin{adjustbox}{width=1\textwidth}
\begin{tabular}{lcccccc}
\hline
\multicolumn{1}{c}{Model}  & Loss   & Top-1   & Top-3   & Top-5   & $R$ & $\#p$      \\ \hline \hline

\multicolumn{1}{l}{ResNet-18}                    
& \multicolumn{1}{l}{0.3181} 
& \multicolumn{1}{l}{92.81} 
& \multicolumn{1}{l}{98.78} 
& \multicolumn{1}{l}{99.49}
& \multicolumn{1}{l}{100}   
& \multicolumn{1}{l}{11.2M}\\

\multicolumn{1}{l}{ResNet-18+DeepCompression}         
& \multicolumn{1}{l}{0.6893} 
& \multicolumn{1}{l}{76.18}                           
& \multicolumn{1}{l}{94.21} 
& \multicolumn{1}{l}{98.63} 
& \multicolumn{1}{l}{49.19} 
& \multicolumn{1}{l}{5.5M}   \\

\multicolumn{1}{l}{ResNet-18+IPruning(F)}        
& \multicolumn{1}{l}{0.5167} 
& \multicolumn{1}{l}{84.12} 
& \multicolumn{1}{l}{96.74} 
& \multicolumn{1}{l}{99.24} 
& \multicolumn{1}{l}{100}  
& \multicolumn{1}{l}{11.2M}\\

\multicolumn{1}{l}{ResNet-18+IPruning(P)}        
& \multicolumn{1}{l}{0.5254} 
& \multicolumn{1}{l}{84.09} 
& \multicolumn{1}{l}{96.77} 
& \multicolumn{1}{l}{99.33} 
& \multicolumn{1}{l}{49.19} 
& \multicolumn{1}{l}{5.5M}\\ \hline\hline

\multicolumn{1}{l}{ResNet-34}                    
& \multicolumn{1}{l}{0.3684} 
& \multicolumn{1}{l}{92.80} 
& \multicolumn{1}{l}{98.85} 
& \multicolumn{1}{l}{99.71} 
& \multicolumn{1}{l}{100}  
& \multicolumn{1}{l}{21.3M}  \\

\multicolumn{1}{l}{ResNet-34+DeepCompression}         
& \multicolumn{1}{l}{0.8423} 
& \multicolumn{1}{l}{71.45}                           
& \multicolumn{1}{l}{93.28} 
& \multicolumn{1}{l}{98.39} 
& \multicolumn{1}{l}{49.61}   
& \multicolumn{1}{l}{10.5M}\\

\multicolumn{1}{l}{ResNet-34+IPruning(F)}         
& \multicolumn{1}{l}{0.6352} 
& \multicolumn{1}{l}{88.78} 
& \multicolumn{1}{l}{98.14} 
& \multicolumn{1}{l}{99.41} 
& \multicolumn{1}{l}{100} 
& \multicolumn{1}{l}{21.3M}  \\

\multicolumn{1}{l}{ResNet-34+IPruning(P)}        
& \multicolumn{1}{l}{0.6401} 
& \multicolumn{1}{l}{88.72}   
& \multicolumn{1}{l}{97.93}   
& \multicolumn{1}{l}{99.42}   
& \multicolumn{1}{l}{49.61} 
& \multicolumn{1}{l}{10.5M} \\ \hline\hline

\multicolumn{1}{l}{ResNet-50}                    
& \multicolumn{1}{l}{0.3761} 
& \multicolumn{1}{l}{92.21} 
& \multicolumn{1}{l}{98.70} 
& \multicolumn{1}{l}{99.51} 
& \multicolumn{1}{l}{100}   
& \multicolumn{1}{l}{23.5M} \\

\multicolumn{1}{l}{ResNet-50+DeepCompression}         
& \multicolumn{1}{l}{1.0355} 
& \multicolumn{1}{l}{67.47}                           
& \multicolumn{1}{l}{90.45} 
& \multicolumn{1}{l}{97.26} 
& \multicolumn{1}{l}{43.46}   
& \multicolumn{1}{l}{10.2M} \\ 

\multicolumn{1}{l}{ResNet-50+IPruning(F)}       
& \multicolumn{1}{l}{0.8200} 
& \multicolumn{1}{l}{82.32} 
& \multicolumn{1}{l}{95.92} 
& \multicolumn{1}{l}{97.37} 
& \multicolumn{1}{l}{100} 
& \multicolumn{1}{l}{23.5M}   \\

\multicolumn{1}{l}{ResNet-50+IPruning(P)}        
& \multicolumn{1}{l}{0.8374} 
& \multicolumn{1}{l}{82.45}   
& \multicolumn{1}{l}{95.32} 
& \multicolumn{1}{l}{97.27} 
& \multicolumn{1}{l}{43.46} 
& \multicolumn{1}{l}{10.2M} \\ \hline\hline

\multicolumn{1}{l}{ResNet-101}                  
& \multicolumn{1}{l}{0.3680} 
& \multicolumn{1}{l}{92.66} 
& \multicolumn{1}{l}{98.69} 
& \multicolumn{1}{l}{99.65} 
& \multicolumn{1}{l}{100}  
& \multicolumn{1}{l}{42.5M}   \\

\multicolumn{1}{l}{ResNet-101+DeepCompression}         
& \multicolumn{1}{l}{1.083} 
& \multicolumn{1}{l}{66.63}                           
& \multicolumn{1}{l}{92.03} 
& \multicolumn{1}{l}{97.97} 
& \multicolumn{1}{l}{42.41}   
& \multicolumn{1}{l}{18.0M} \\ 

\multicolumn{1}{l}{ResNet-101+IPruning(F)}       
& \multicolumn{1}{l}{0.8233} 
& \multicolumn{1}{l}{84.47} 
& \multicolumn{1}{l}{97.42} 
& \multicolumn{1}{l}{98.47} 
& \multicolumn{1}{l}{100}  
& \multicolumn{1}{l}{42.5M}   \\ 

\multicolumn{1}{l}{ResNet-101+IPruning(P)}       
& \multicolumn{1}{l}{0.8372} 
& \multicolumn{1}{l}{84.38} 
& \multicolumn{1}{l}{97.03} 
& \multicolumn{1}{l}{98.37} 
& \multicolumn{1}{l}{42.41} 
& \multicolumn{1}{l}{18.0M} \\ \hline\hline

\multicolumn{1}{l}{AlexNet}       
& \multicolumn{1}{l}{0.9727} 
& \multicolumn{1}{l}{84.32} 
& \multicolumn{1}{l}{96.58} 
& \multicolumn{1}{l}{99.08} 
& \multicolumn{1}{l}{100} 
& \multicolumn{1}{l}{57.4M} \\

\multicolumn{1}{l}{AlexNet+IPruning(F)}       
& \multicolumn{1}{l}{0.8842} 
& \multicolumn{1}{l}{74.02} 
& \multicolumn{1}{l}{92.79} 
& \multicolumn{1}{l}{97.63} 
& \multicolumn{1}{l}{100} 
& \multicolumn{1}{l}{57.4M} \\

\multicolumn{1}{l}{AlexNet+IPruning(P)}       
& \multicolumn{1}{l}{0.8830} 
& \multicolumn{1}{l}{73.62} 
& \multicolumn{1}{l}{92.35} 
& \multicolumn{1}{l}{97.03} 
& \multicolumn{1}{l}{62.84} 
& \multicolumn{1}{l}{36.0M} \\

\multicolumn{1}{l}{SqueezeNet}       
& \multicolumn{1}{l}{0.5585} 
& \multicolumn{1}{l}{81.49} 
& \multicolumn{1}{l}{96.31} 
& \multicolumn{1}{l}{99.01} 
& \multicolumn{1}{l}{100} 
& \multicolumn{1}{l}{0.73M} \\

\multicolumn{1}{l}{SqueezeNet+IPruning(F)}       
& \multicolumn{1}{l}{0.6894} 
& \multicolumn{1}{l}{76.74} 
& \multicolumn{1}{l}{95.53} 
& \multicolumn{1}{l}{98.54} 
& \multicolumn{1}{l}{100} 
& \multicolumn{1}{l}{0.73M} \\

\multicolumn{1}{l}{SqueezeNet+IPruning(P)}       
& \multicolumn{1}{l}{0.6989} 
& \multicolumn{1}{l}{76.35} 
& \multicolumn{1}{l}{95.13} 
& \multicolumn{1}{l}{98.34} 
& \multicolumn{1}{l}{51.26}
& \multicolumn{1}{l}{0.37M} \\  \hline
\end{tabular}
\end{adjustbox}
\label{T:results_cifar10_IPruning}
\end{subtable}

\begin{subtable}{0.99\linewidth}
\centering
\captionsetup{font=footnotesize}
\caption{\textbf{CIFAR-100}}
\begin{adjustbox}{width=1\textwidth}
\begin{tabular}{lcccccc}
\hline
\multicolumn{1}{c}{Model}  & Loss   & Top-1   & Top-3   & Top-5   & $R$ & $\#p$      \\ \hline \hline

\multicolumn{1}{l}{ResNet-18}                  
& \multicolumn{1}{l}{1.3830}
& \multicolumn{1}{l}{69.03} 
& \multicolumn{1}{l}{84.44} 
& \multicolumn{1}{l}{88.90} 
& \multicolumn{1}{c}{100}   
& \multicolumn{1}{l}{11.2M}\\

\multicolumn{1}{l}{ResNet-18+DeepCompression}         
& \multicolumn{1}{l}{2.2130} 
& \multicolumn{1}{l}{40.15}                           
& \multicolumn{1}{l}{61.92} 
& \multicolumn{1}{l}{71.84} 
& \multicolumn{1}{l}{47.95}  
& \multicolumn{1}{l}{5.3M} \\ 

\multicolumn{1}{l}{ResNet-18+IPruning(F)}        
& \multicolumn{1}{l}{1.8431}
& \multicolumn{1}{l}{55.43} 
& \multicolumn{1}{l}{74.94}   
& \multicolumn{1}{l}{82.60}   
& \multicolumn{1}{c}{100}   
& \multicolumn{1}{l}{11.2M}\\

\multicolumn{1}{l}{ResNet-18+IPruning(P)}    
& \multicolumn{1}{l}{1.8696} 
& \multicolumn{1}{l}{56.43} 
& \multicolumn{1}{l}{75.37} 
& \multicolumn{1}{l}{82.43}   
& \multicolumn{1}{c}{47.95}
& \multicolumn{1}{l}{5.3M}\\  \hline\hline

\multicolumn{1}{l}{ResNet-34}                  
& \multicolumn{1}{l}{1.3931} 
& \multicolumn{1}{l}{69.96} 
& \multicolumn{1}{l}{85.65}  
& \multicolumn{1}{l}{90.10} 
& \multicolumn{1}{l}{100}  
& \multicolumn{1}{l}{21.3M}  \\

\multicolumn{1}{l}{ResNet-34+DeepCompression}         
& \multicolumn{1}{l}{2.1778} 
& \multicolumn{1}{l}{42.09} 
& \multicolumn{1}{l}{65.01}                           
& \multicolumn{1}{l}{74.31} 
& \multicolumn{1}{l}{49.41} 
& \multicolumn{1}{l}{10.5M}   \\

\multicolumn{1}{l}{ResNet-34+IPruning(F)}     
& \multicolumn{1}{l}{2.3789} 
& \multicolumn{1}{l}{60.73}  
& \multicolumn{1}{l}{79.26}   
& \multicolumn{1}{l}{85.48}   
& \multicolumn{1}{l}{100}   
& \multicolumn{1}{l}{21.3M}  \\

\multicolumn{1}{l}{ResNet-34+IPruning(P)}       
& \multicolumn{1}{l}{2.3794} 
& \multicolumn{1}{l}{61.13}  
& \multicolumn{1}{l}{79.23}  
& \multicolumn{1}{l}{85.30}   
& \multicolumn{1}{c}{49.41}  
& \multicolumn{1}{l}{10.5M}  \\ \hline\hline

\multicolumn{1}{l}{ResNet-50}                    
& \multicolumn{1}{l}{1.3068}
& \multicolumn{1}{l}{71.22}  
& \multicolumn{1}{l}{86.47}  
& \multicolumn{1}{l}{90.74}  
& \multicolumn{1}{c}{100}  
& \multicolumn{1}{l}{23.7M}  \\

\multicolumn{1}{l}{ResNet-50+DeepCompression}         
& \multicolumn{1}{l}{2.4927} 
& \multicolumn{1}{l}{43.72}                           
& \multicolumn{1}{l}{66.93} 
& \multicolumn{1}{l}{76.15} 
& \multicolumn{1}{l}{44.63}  
& \multicolumn{1}{l}{10.8M}  \\

\multicolumn{1}{l}{ResNet-50+IPruning(F)}        
& \multicolumn{1}{l}{1.8750} 
& \multicolumn{1}{l}{60.44}   
& \multicolumn{1}{l}{79.25}   
& \multicolumn{1}{l}{86.24}   
& \multicolumn{1}{c}{100}  
& \multicolumn{1}{l}{23.7M}  \\ 

\multicolumn{1}{l}{ResNet-50+IPruning(P)}    
& \multicolumn{1}{l}{2.1462} 
& \multicolumn{1}{l}{60.05}  
& \multicolumn{1}{l}{78.83}  
& \multicolumn{1}{l}{85.78} 
& \multicolumn{1}{c}{44.63} 
& \multicolumn{1}{l}{10.8M}  \\ \hline\hline 

\multicolumn{1}{l}{ResNet-101}                   
& \multicolumn{1}{l}{1.3574} 
& \multicolumn{1}{l}{71.19}   
& \multicolumn{1}{l}{85.54}   
& \multicolumn{1}{l}{90.00}   
& \multicolumn{1}{c}{100}  
& \multicolumn{1}{c}{42.6M}   \\

\multicolumn{1}{l}{ResNet-101+DeepCompression}         
& \multicolumn{1}{l}{2.6232} 
& \multicolumn{1}{l}{36.58}                           
& \multicolumn{1}{l}{57.82} 
& \multicolumn{1}{l}{68.36} 
& \multicolumn{1}{l}{41.36}  
& \multicolumn{1}{c}{17.6M}   \\

\multicolumn{1}{l}{ResNet-101+IPruning(F)}       
& \multicolumn{1}{l}{2.1338} 
& \multicolumn{1}{l}{60.52}   
& \multicolumn{1}{l}{79.91}   
& \multicolumn{1}{l}{83.22}   
& \multicolumn{1}{c}{100} 
& \multicolumn{1}{c}{42.6M}   \\

\multicolumn{1}{l}{ResNet-101+IPruning(P)}       
& \multicolumn{1}{l}{2.2952} 
& \multicolumn{1}{l}{60.35} 
& \multicolumn{1}{l}{78.99} 
& \multicolumn{1}{l}{83.01} 
& \multicolumn{1}{c}{41.36} 
& \multicolumn{1}{c}{17.6M}   \\ \hline\hline

\multicolumn{1}{l}{AlexNet}       
& \multicolumn{1}{l}{2.8113} 
& \multicolumn{1}{l}{60.12} 
& \multicolumn{1}{l}{79.18} 
& \multicolumn{1}{l}{83.31} 
& \multicolumn{1}{l}{100} 
& \multicolumn{1}{l}{57.4M} \\

\multicolumn{1}{l}{AlexNet+IPruning(F)}       
& \multicolumn{1}{l}{2.7420} 
& \multicolumn{1}{l}{53.52} 
& \multicolumn{1}{l}{72.42} 
& \multicolumn{1}{l}{79.70} 
& \multicolumn{1}{l}{100} 
& \multicolumn{1}{l}{57.4M} \\

\multicolumn{1}{l}{AlexNet+IPruning(P)}       
& \multicolumn{1}{l}{2.7396} 
& \multicolumn{1}{l}{53.05} 
& \multicolumn{1}{l}{72.28} 
& \multicolumn{1}{l}{79.69} 
& \multicolumn{1}{l}{65.35} 
& \multicolumn{1}{l}{37.5M} \\

\multicolumn{1}{l}{SqueezeNet}       
& \multicolumn{1}{l}{1.4150} 
& \multicolumn{1}{l}{67.85} 
& \multicolumn{1}{l}{85.81} 
& \multicolumn{1}{l}{89.69} 
& \multicolumn{1}{l}{100}
& \multicolumn{1}{l}{0.77M} \\

\multicolumn{1}{l}{SqueezeNet+IPruning(F)}       
& \multicolumn{1}{l}{1.9285} 
& \multicolumn{1}{l}{61.93} 
& \multicolumn{1}{l}{80.74} 
& \multicolumn{1}{l}{86.92} 
& \multicolumn{1}{l}{100}
& \multicolumn{1}{l}{0.77M} \\

\multicolumn{1}{l}{SqueezeNet+IPruning(P)}       
& \multicolumn{1}{l}{1.9437} 
& \multicolumn{1}{l}{61.46} 
& \multicolumn{1}{l}{80.45} 
& \multicolumn{1}{l}{85.81} 
& \multicolumn{1}{l}{53.20} 
& \multicolumn{1}{l}{0.41M} \\ \hline

\end{tabular}
\end{adjustbox}
\label{T:results_cifar100_IPruning}
\end{subtable}

\label{T:results_IPruning}
\vspace{-6mm}
\end{table}

\section{Experiments}
\label{sec:experiemnts}
The experiments were conducted on the CIFAR-10~and 
CIFAR-100~\cite{krizhevsky2009learning} datasets using ResNets (18, 34, 50, and 101 layers)~\cite{he2016deep}, AlexNet~\cite{krizhevsky2012imagenet}, SqueezeNet~\cite{iandola2016squeezenet}, and Deep Compression~\cite{han2015deep}. Horizontal flip and Cutout~\cite{devries2017improved} augmentation methods were used.
The results are averaged over five independent runs. The Adadelta optimizer with Step adaptive learning rate (step: every 50 epoch at gamma rate of 0.1) and weight decay of $10e^{-6}$ is used. The number of epochs is 200 and the batch size is 128. Random dropout rate is set to 0.5 where applicable, except for the proposed model. The early state convergence in~(\ref{eq:stateconvergence}) is used with a threshold of 100. 

As Table~\ref{T:results_IPruning} shows, IPruning on average has removed more than $50\%$ of the trainable weights and the Top-1 performance has dropped less than $10\%$ compared to the original model. We used the pruning rate achieved by IPruning to prune the original network using Deep Compression \cite{han2015deep}. Since this method is tailored to pruning certain layers, we have modified it to prune every layer, similar to IPruning. We also have evaluated inference results of IPruning in full and pruned modes. The former refers to training the network with IPruning but performing inference using the full model, and the latter refers to training the network with IPruning and performing inference with the pruned network. The results show that the full network has slightly better performance than the pruned network. It shows that we are able to achieve very competitive performance using the pruned network compared with the full network, which has a larger capacity, trained with IPruning.

\begin{figure}[!t]
\centering
\centering
\captionsetup{font=footnotesize}
\begin{subfigure}[t]{0.4\textwidth}
\captionsetup{font=footnotesize}
\centering
\includegraphics[width=1\textwidth]{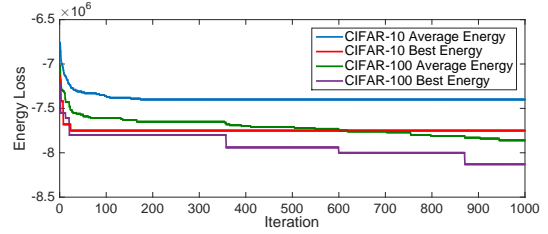}
\caption{Average energy and best energy of states population.}
\label{fig:}
\end{subfigure}%

\begin{subfigure}[t]{0.4\textwidth}
\captionsetup{font=footnotesize}
\centering
\includegraphics[width=1\textwidth]{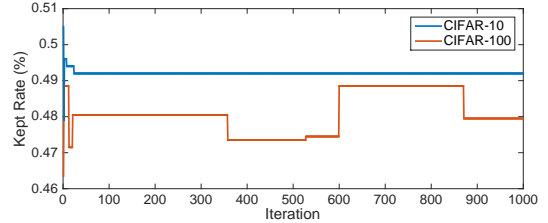}  
\caption{Rate of kept trainable parameters.}
\label{fig:}
\end{subfigure}%
\caption{Energy loss and kept rate for ResNet-18 over 1,000 training iterations.}
\vspace{-6mm}
\label{fig:visualization_cifar_res18_idropout}
\end{figure}

Figure~\ref{fig:visualization_cifar_res18_idropout} shows the energy loss and corresponding pruning rate over 1,000 training iterations of IPruning for ResNet-18. Since CIFAR-100 is more complicated than CIFAR-10, it converges slower. Results show that the pruning rates generally converge to a value close to $50\%$, regardless of the initial distribution of the population. This might be due to the behavior of optimizer in very high dimensional space and its limited capability of reaching all possible states during the evolution.

\section{Conclusions}
\label{sec:conclusion}
We propose an Ising energy-based framework, called IPruning, for structured pruning of neural networks. Unlike most other methods, IPruning considers every trainable weight in any layer of a given network for pruning. From an implementation perspective, most pruning methods require manual modification of network architecture to apply the pruning mask while IPruning can automatically detect trainable weights and construct a pruning graph for a given network.

\section{Acknowledgment}
The authors acknowledge financial support of Fujitsu Laboratories Ltd. and Fujitsu Consulting (Canada) Inc.
\bibliographystyle{IEEEbib}
\bibliography{strings,mybibfile}

\end{document}